# Life-Long Multi-Task Learning of Adaptive Path Tracking Policy for Autonomous Vehicle

Cheng Gong, *Student Member IEEE*, Jianwei Gong, *Member IEEE*, Chao Lu, Zhe Liu, Zirui Li

*Abstract—* This paper proposes a life-long adaptive path tracking policy learning method for autonomous vehicles that can self-evolve and self-adapt with multi-task knowledge. Firstly, the proposed method can learn a model-free control policy for path tracking directly from the historical driving experience, where the property of vehicle dynamics and corresponding control strategy can be learned simultaneously. Secondly, by utilizing the life-long learning method, the proposed method can learn the policy with task-incremental knowledge without encountering catastrophic forgetting. Thus, with continual multi-task knowledge learned, the policy can iteratively adapt to new tasks and improve its performance with knowledge from new tasks. Thirdly, a memory evaluation and updating method is applied to optimize memory structure for life-long learning which enables the policy to learn toward selected directions. Experiments are conducted using a high-fidelity vehicle dynamic model in a complex curvy road to evaluate the performance of the proposed method. Results show that the proposed method can effectively evolve with continual multi-task knowledge and adapt to the new environment, where the performance of the proposed method can also surpass two commonly used baseline methods after evolving.

## I. INTRODUCTION

Accurate path tracking control for autonomous vehicles (AV) to follow the desired path is an essential technology for guarantying the safety, stability and riding comfort in autonomous driving. Traditional path tracking methods are mostly based on static linear models or massive manual tuning of algorithm parameters using expert prior knowledge, including PID control[1], feedback-feedforward control[2], optimal control[3-5], and etc. [6, 7]. These methods can work well under designed working conditions, but are usually sensitive to changes in working conditions or model parameters, which lead to their poor performance in actively adapting to different vehicles, driving tasks, and driving environments.

In terms of improving the adaptability of the path tracking control methods, intelligent control methods have shown great superiority in model recognition and parameter adjusting. Based on expert knowledge and prior experience, many fuzzy-based methods [8-12] and adaptive-laws [13] are proposed to better model the vehicle dynamic model. But the accuracy of these methods greatly depends on the properly modelling of the fuzzy logic and adaptive rules. To model complex or unknown dynamics more precisely, machine learning methods can be used to learn to vehicle and environmental property using posterior knowledge [14-17]. [14] employs an ANN to approximate tire cornering stiffness. [18] employed a neural network to update the dynamic model for skid-steered robot. [16, 17] integrates Gaussian Mixture Model and Gaussian Mixture Regression with pure pursuit method in constructing personalized path tracking policy. However, these methods can only learn the model partially and locally, and still rely on explicit model representation, where the effort for fine-tuning model parameters and adjusting control schemes to adapt to different vehicles and environments are inevitable.

Instead of constantly fine-tuning or modifying the traditional methods for model adapting, many researchers also seek to improve the model adaptability through learning driving experience directly through imitate learning [19-23] or reinforcement learning [24-26]. By exploiting and imitating the collected posterior driving experience, these methods can learn the vehicle dynamics and control policy adaptively and avoid complex model approximating and parameter tuning. But there are still some critical issues preventing them from the real-world application, where one major issue is the lack of online adaptability to continually learning multi-tasks. Although reinforcement-learning-based methods can fully explore action space and learn an approximate policy, the policy exploring can be very time-consuming, which prohibits the online adapting of the policy. Meanwhile, imitate learning methods are ideal for online policy adaptation and improvement, as they learn policy directly from historical experience. With the accumulation of task knowledge, the policy is expected to evolve and generalize to different conditions. But catastrophic forgetting may happen in the process, where previously learned knowledge may be forgotten when the network is learning on new knowledge.

The catastrophic forgetting issue has troubled researcher for many years, and many methods were proposed to moderate this issue, namely, life-long learning methods or incremental learning methods [27]. The goal of life-long learning is to learn knowledge in new tasks without forgetting the learned knowledge in previous tasks [28]. This principle of life-long learning makes it very applicable for robots, where robots are expected to continually exploring new environments and learning to adapt to new environments while not forgetting knowledge in preview tasks. An early attempt in robot navigation is introduced in [29], and a recent attempt is presented in [30] where the life-long learning method is utilized to lean the end-to-end robot navigation policy. Recently, life-long learning methods are also applied in formulating continual behavior prediction for multi-agent interaction [31]. However, to the author's knowledge, few researches have paid attention to exploring the possibility of continual control policy learning through life-long learning, which can enable the policy to fully exploit the experience and improve with task experience accumulated.

* Corresponding author: Jianwei Gong.
The authors are with the Research Center of Intelligent Vehicle, School of Mechanical Engineering, Beijing Institute of Technology, Beijing 100081, China. (chenggong@bit.edu.cn; gongjianwei@bit.edu.cn)
Zirui Li is also with the Department of Transport and Planning, Faculty of Civil Engineering and Geosciences, Delft University of Technology, Stevinweg 1, 2628 CN Delft, The Netherlands.

In this paper, an adaptive path tracking policy learning method for continually multi-task learning is proposed based on life-long learning. The proposed method learns the policy on processed historical driving experience directly in a model-free manner, where the real vehicle dynamic property and corresponding control policy can be learned simultaneously. With the life-long learning method utilized, the method can learn on continual task knowledge to improve its performance and adaptivity in more complex environments without encountering catastrophic forgetting. Experiments in MATLAB/SIMULINK using high fidelity vehicle dynamics are conducted to demonstrate the validity and effectiveness of the proposed method. The main contributions of this paper are as follows:

- A life-long learning framework of policy learning for path tracking is proposed, which can gradually improve the performance of the policy and generalize it to different scenarios by learning the continual multi-task knowledge collected from applications without encountering catastrophic forgetting.

- A memory evaluation and updating method for optimizing memory structure of life-long policy learning is proposed, which can further help the path tracking policy in selectively learning desired features to improve the policy performance.

- A novel model-free path tracking policy is proposed that can learn vehicle dynamic responses and corresponding control action simultaneously from historical driving data.

## II. PROBLEM FORMULATION

In this section, the formulation of the path tracking problem will be introduced first, where the path tracking task will be defined and the corresponding path tracking policy will be introduced. Then, the mechanism of the life-long learning method used in this study is described, which uses episodic memory to constraint learning direction and avoid catastrophic forgetting.

### A. Path-Tracking Problem

In the problem of path-tracking, the vehicle needs to properly control the vehicle steering motion so as to track the desired path. The performance of a path-tracking policy can be evaluated through several indicators, for instance, the lateral tracking error, the control effort, and riding comfort. The illustration of a path-tracking process is shown in Fig.1.

To better understand and learn the inherent control policy from experience, the path-tracking process needs to be properly modeled and represented. An explicit and explainable representation is necessary to guarantee scene understanding and learning performance, where features should be carefully selected and described. In this paper, to model the path tracking process, two basic aspects of knowledge are considered. The first aspect is the information of the path to be tracked, which is usually a continuous curve and should be properly described to form the goal of the policy. Several common methods can be used to represent the reference path such as discretized position set, the discretized curvature set, the preview point, and diagrams of the path. Another aspect is the understanding of the vehicle dynamic model property, which determines how the vehicle will respond to the control action. The vehicle dynamic property is usually vehicle-specific and relies on a large number of factors and vehicle parameters, among which many are also time-variant or highly nonlinear.

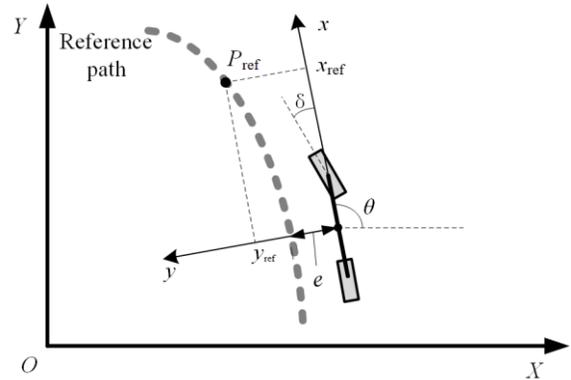

Figure 1. Path tracking scenario with simplified bicycle model.

For the first aspect, a preview point scheme similar to the pure-pursuit method is applied in this study, where the reference path is represented by a preview point up ahead. And the preview point is determined from the reference path using a specific lookahead distance from the vehicle. By applying the preview point for representation, the continuous reference curve can be decomposed to one simple position per control period, which is a rather convenient and effective way to represent the reference path. For the aspect of vehicle dynamic property representation, the vehicle dynamic property will be determined using direct vehicle dynamic status, such as velocity. And thus, combining the knowledge of the above two aspects, the policy $\pi$ for vehicle path tracking can be formulated as:

$$\delta = \pi(P_{\text{ref}}, \xi), \quad (1)$$

where $P_{\text{ref}} = [x_{\text{ref}}, y_{\text{ref}}]$ is the preview point, $\xi = [v_x, v_y, \dot{\theta}]$ represents the vehicle dynamic status profile.

### B. Average Gradient Episodic Memory

Average Gradient Episodic Memory (AGEM) [32] is a life-long learning algorithm invented to avoid forgetting old knowledge from previous tasks when learning new knowledge in new tasks. It is an improved version of Gradient Episodic Memory (GEM) [33], and is more computationally efficient, which makes it more applicable to applications that demand faster training speed and lower computational burden. To avoid catastrophic forgetting, AGEM maintains an episodic memory, which stores knowledge from previews tasks. The episodic memory will be used to compute episodic memory loss when learning new knowledge, and AGEM prevents forgetting by decreasing the training loss on the new task while constraining episodic memory loss from increasing. Instead of computing the episodic memory loss of every previous task in the training process, AGEM computes the average episodic memory loss, and the object of AGEM in learning a new task $t$ can be expressed as:

$$\min_{\theta} l(f_\theta, \mathcal{D}_t) \quad s.t. \quad l(f_\theta, \mathcal{D}_t) \leq l(f_\theta^{t-1}, \mathcal{M}), \quad (2)$$

where the $f_\theta$ is the learning model with parameter $\theta$, $\mathcal{D}_t$ is the data acquired in the new task $t$, and $f_\theta^{t-1}$ is the model trained till task $t-1$, $\mathcal{M} = \cup_{k<t} \mathcal{M}_k$ is a random sampled batch of the episodic memory of previous tasks in which $\mathcal{M}_k$ is the episodic memory of the $k_{\text{th}}$ task. The optimization of the loss is then reduced to the optimization of the model gradients, where the gradient for decreasing training loss should have the same direction as the gradient for decreasing episodic memory loss:

$$\min_{\tilde{g}} \|\tilde{g} - g\|_2^2 \quad s.t. \quad \tilde{g}^\mathrm{T} g_{\text{ref}} \geq 0, \quad (3)$$

where $g$ is the gradient calculated in training the current task, and $g_{\text{ref}}$ is the reference gradient calculated using the random sampled batch of the episodic memory $\mathcal{M}$. Compared to GEM, which solves the optimization problem in (3) through quadratic program (QP), AGEM proposed a more effective solution, which derives the solution when directions of two gradients contradict via:

$$\tilde{g} = g - \frac{g^\mathrm{T} g_{\text{ref}}}{g_{\text{ref}}^\mathrm{T} g_{\text{ref}}} g_{\text{ref}}, \quad (4)$$

where the gradient can be directly computed and is very time-efficient compared to solve a QP problem.

### III. LIFE-LONG LEARNING FOR PATH-TRACKING

Ideally, similar to human drivers that can learn to drive better with driving experience increases, the policy is expected to be iteratively evolving and improving with more and more knowledge acquired, and eventually verge on optimal performance. However, direct policy imitating from human behaviors or other controllers may result in involving erratic behavior and learning the wrong demonstration that undermines the learned policy [34]. Besides, catastrophic forgetting may happen in the process of learning new knowledge where the learned policy may forget the learned knowledge from previous tasks. And thus, the life-long policy learning framework is proposed to properly learn the path tracking policy while avoiding catastrophic forgetting and improve the performance of the learned policy.

Instead of imitating the control policy directly from the experience of path tracking control, we can fine-tune the knowledge with realistic vehicle response and reversibly learn the corresponding vehicle control policy. Since the policy learns from collected driving experience, the historical control action and corresponding vehicle trajectory can be acquired as the posterior knowledge. We can regard that the vehicle's historical trajectory as the reference path and hence the experience can be regarded as perfectly path tracking knowledge without tracking error. Thus, the realistic control error of the path tracking knowledge can be excluded in policy learning, and realistic vehicle responses and corresponding actions can be learned. And the knowledge processed in task t can be described as:

$$\mathcal{D}_t = \{(s_k, a_k) | k = 1, \ldots, N\} \quad (5)$$

where $s_k = [x_{\text{ref}}, y_{\text{ref}}, v_x, v_y, v_r]$ represents the model states which include the vehicle dynamic status and path tracking goal, and $a_k = \delta_k$ represents the corresponding control action. The training loss of the policy is measured by mean square error (MSE):

$$l(\pi_\theta, \mathcal{D}_t) = \frac{1}{N} \sum_{k=1}^{N} (\pi_\theta(s_k) - a_k)^2, \quad (s_k, a_k) \in \mathcal{D}_t \quad (6)$$

To enable the policy to evolve with more task experience accumulated without forgetting prior task knowledge, the AGEM is utilized in learning the path tracking policy in a continual multi-task manner, which is achieved through constraining that the decrease of the loss in the new task does not increase the loss in prior tasks:

$$\min_{\theta} l(\pi_\theta, \mathcal{D}_t) \quad s.t. \quad l(\pi_\theta, \mathcal{D}_t) \leq l(\pi_\theta^{t-1}, \mathcal{M}), \quad (7)$$

where the goal is to learn the path-tracking policy $\pi_\theta$ from the data collected in task $t$ while not degrading its performance in the episodic memory collected from previous tasks. For autonomous systems, different tasks usually refer to missions in different environments or with different goals. And thus, in this problem, the tasks can be determined by different road geometry information and different vehicle dynamic statuses. Different tasks are then separated from collected data into time-continuous sets using different road sections and different tracking velocities as indicators, which is very convenient for onboard online policy learning and updating.

By applying AGEM, the catastrophic forgetting problem can be moderated when learning new knowledge from new tasks. But in the path tracking problem, it is not guaranteed that the knowledge from the new task will always surpass the prior acquainted knowledge. Since the episodic memory is served as the constraint of the learning direction, we will consider a different memory updating rule. On the one hand, the policy is expected to be learned to generalize to more different environments. On the other hand, we want the learned policy to improve its performance in old-task while learning new knowledge in new-task. Thus, the memory is constructed to be equally distributed and cover the state space as much as possible, while the performance of memory should be compared with new knowledge and always retain the one with the best performance. To measure the similarity between different data, Euclidian distance will be calculated between new data and data within the episodic memory. Data from a newly learned task knowledge will be added if:

$$sim(s_k, s_j) = \|s_k - s_j\|_2^2 > \eta, \quad (s_k, a_k) \in D_t, \forall (s_j, a_j) \in \mathcal{M} \quad (8)$$

where $\eta$ is the similarity threshold to ensure the knowledge is evenly distributed in episodic memory, and can be adjusted according to desired episodic memory. In some case, the knowledge from new task may contradict with the prior knowledge stored in episodic memory, and the evaluation will be conducted and the knowledge with better evaluation score will be stored:

$$(s_o, a_o) = \arg\min_{(s_j, a_j) \in S_k} EVAL(s_j, a_j), \quad (9)$$

where $S_k$ is the set that stores similar knowledge corresponding to $(s_k, a_k) \in D_t$:

$$S_k = \{(s_k, a_k) \cup (s_j, a_j) | sim(s_k, s_j) \leq \eta, (s_j, a_j) \in M\}, \quad (10)$$

where $(s_o, a_o)$ is the optimal knowledge in $S_k$ and will be stored to episodic memory while other memory in $S_k$ will be excluded from episodic memory $\mathcal{M}$.

## IV. EXPERIMENT

To examine and analyze the validity of the proposed method, we will introduce the experiments as well as the experimental results in this section. First, the experimental environment and its setup will be described, where both the data collection and policy evaluation are conducted. Then the results of policy training are presented to statistically illustrate the importance of using life-long learning for continual multi-task policy learning. Lastly, the learned policies will be examined and evaluated in the test scene where their performance will also be compared with two baseline methods.

### A. Experimental Environment and Data Collection

To acquire the data for policy training as well as policy evaluation, a simulation environment that can approximate complex vehicle dynamics is built base on MATLAB/SIMULINK and Vehicle Dynamic Toolbox [35]. The illustration of the experimental environment is shown in Fig.2, where a vehicle dual-track model with 3 DOF is adopted for vehicle simulation, and a curved road pre-set scenario is selected as the experimental scenario which contains rich road characteristics with various road curvatures. The reference path is extracted based on interpolation of a set of selected waypoints that are manually selected from the curved road. Three road sections with different characteristics were segmented from the reference path to examine the adaptability of the learned policy and avoid overfitting.

To collect data for policy learning, model predictive control (MPC) based on the vehicle dynamic model are applied to track the path and provide driving data. To gather as much driving information as possible, path tracking is conducted with different velocity profiles in each section several times. Since no longitudinal speed planning is considered in this research while the tracking velocity varies from 3m/s to 15m/s, the baseline method may fail to perform path tracking in a sharp turn and the failed data will be neglected from the collected dataset. A neural network with two hidden layers and 64 units per layer is applied for learning policy $\pi_\theta$. The lookahead distance is set to 2m ahead. And the average inference time of the policy is 1.1ms, which is capable of performing in real-time.

### B. Evaluation of Policy Learning Performance

To demonstrate the effectiveness of utilizing the life-long scheme in continual multi-task policy learning without catastrophic forgetting, the policy learning performances of different learning methods in continually learning multi-task knowledge are evaluated and compared. Knowledge from different tasks are collected from different road sections in the experimental environment with different reference velocity, and are applied to different learning methods for sequential multi-task training. Three different learning methods are evaluated and compared:

- Non-life-long learning method (Non-LL): Regular policy learning method using gradient descent without considering the constraint in (7).
- Life-long learning method without considering memory evaluation (LL-no-ME): Life-long learning method that considers the constraint in (7), but does not consider the memory evaluation scheme in (9) where the memory will be randomly sampled instead of evaluating.
- Life-long learning method with memory evaluation (LL-ME): Life-long learning method that considers the constraint in (7) and considers the memory evaluation in (9), where the evaluation is chosen as the minimal steering effort:

$$EVAL(s, a) = \|a\|_2^2 \quad (11)$$

The results are shown in Fig.3, where the 25 groups of collected task knowledge are first divided into training sets and test sets, and training sets are applied in learning the policy with the above three learning methods. While the learning performances of those methods are evaluated by average MSE in test sets of all learned tasks. The average MSE in continually learning till $k^{th}$ task can be calculated as:

$$B_k = \frac{1}{k} \sum_{j=1}^{k} b_{k,j}, \quad (12)$$

where $b_{k,j}$ is the average MSE on the test set of the $j^{th}$ task after leaning on the $k^{th}$ task.

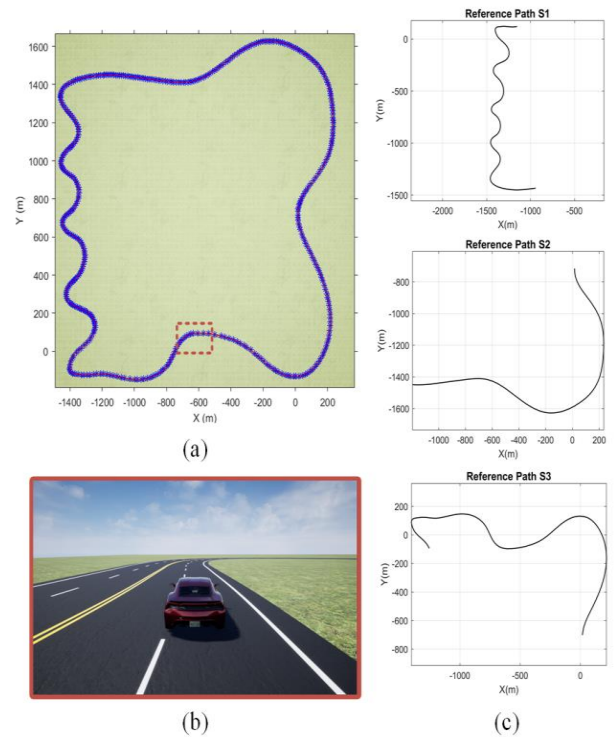

Figure 2. Experimental environment, (a) the experimental pre-set scenario curved road, (b) visualization of the experimental environment, (c) three segmentations of extracted reference paths S1, S2, and S3.

As the results show, the performances of all three methods in all tasks can converge after learning three to four tasks. However, with more new knowledge learned, the Non-LL method may suffer from the catastrophic forgetting issue where the policy may be no longer fit for prior tasks. Compared to LL-no-ME and LL-ME, the Non-LL method performs rather worse after learned the 15$^{th}$ task and 40$^{th}$ task, which is mainly due to the catastrophic forgetting problem. This problem prohibits Non-LL methods from continually learning multi-task knowledge. By utilizing life-long methods, LL-no-ME and LL-ME can learn new knowledge while maintaining the learning performance in all tasks, enabling its application in online algorithm adapting and evolving. But the LL-ME performs slightly more unstable compared to LL-no-ME, where the evaluation and selection of memory may be the may cause.

As Fig.4 shows, the life-long learning method can maintain the policy performance at a relatively stable level after evolved with a certain number of tasks. While the Non-LL scheme may encounter catastrophic forgetting problems even after policy performance has converged. More specifically, as shown in Fig.3 and Fig.4, the Non-LL policy suffers from the increase in Average MSE after learning the 40th task which also leads to the increased deviation in path tracking. Meantime, both average MSE and deviation can be maintained with life-long learning methods with continual task knowledge learned.

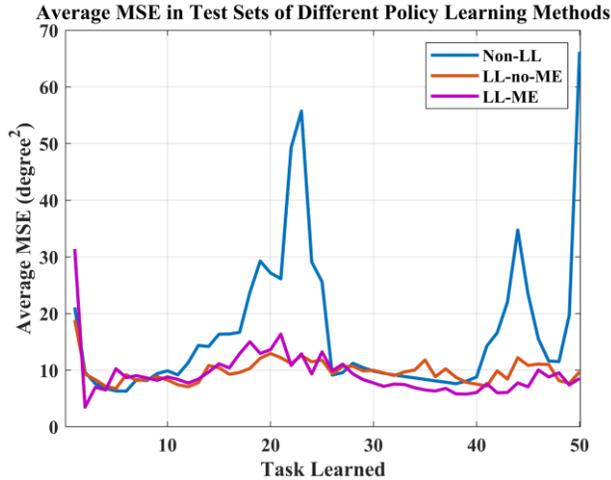

Figure 3.  Learning performance of different learning methods in test sets with multi-task knowledge learned.

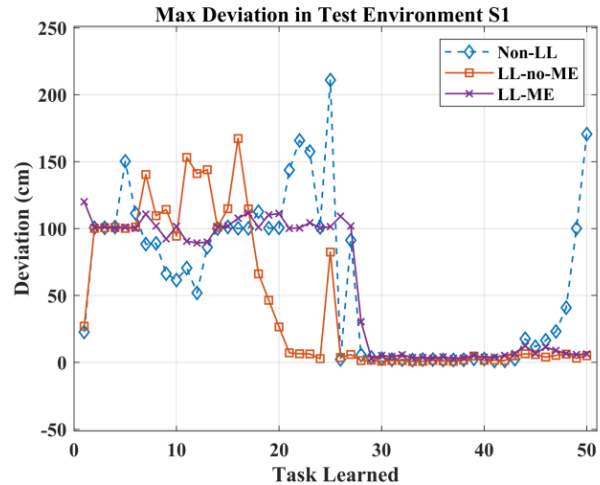

Figure 4.  Max lateral deviation of executing path tracking with three learned policies in test scenario S1.

### C. Policy Evaluation in Path Tracking Task

To Evaluate the applicability of the proposed policy learning method in the path tracking scenario, the policy is iteratively trained and applied in the experimental environment. To avoid overfitting, the training process is conducted in two segmented paths S2 and S3 with different reference velocities, while segmented path S1 is used as the test scenario only. After continually learned on each new task, the policy will be applied in test scenario S1 to perform path tracking with the reference velocity of 10m/s. The results are shown in Fig.4, where performances of three learning methods are shown and compared. The mean and max deviation in path tracking with different policies are used as the performance indicators. All three methods converge only after the learning on the 20$^{th}$ task, which is mainly due to the learning environment S2 and S3 are more simple and easier compared to test environment S1. Although learning on S1 the policy performance can easily converge with experience of fewer tasks learned, it is not our desire to show that the policy learned in difficult tasks can easily be applied to easier tasks. With this experimental result, we want to demonstrate that with continually multi-task knowledge learned, even the accumulation of experience in easy tasks can contribute to the policy application in more difficult tasks.

To further analyze the performance of the proposed method, the performances of learned policies will be analyzed in detail and will be compared with two baseline methods, which are pure pursuit (PP) and the MPC used for data collection (dynamic-MPC). Since the Non-LL method encounters the catastrophic forgetting problem after around the 40$^{th}$ task, where it is no longer capable to fulfill the path-tracking task in test scenario S1, the performances of all methods after learning the 39th task are compared and presented in Fig.5. As the results shown in Fig.5 (a), the learned policies can achieve minor lateral tracking error compared to the two baseline methods. This demonstrates the adaptability of the policy learning methods in learning a more realistic vehicle dynamic model, where the nonlinearity of the dynamic model can be better approximated through learning from the realistic driving experience.

Among the learned policies, the LL-ME achieves the minimum tracking error, where Non-LL can achieve similar performance and LL-no-ME shows greater error. The decrease in tracking error by applying the memory evaluation scheme indicates that the evaluation and moderation of episodic memory can be beneficial for better policy performance. As episodic memory is used as the constraint for policy learning direction, the updating and optimizing of memory structure and quality may help policy to better evolve through continually learning multi-task knowledge. Although Non-LL can also achieve adequate performance before encountering the catastrophic forgetting problem, it may also lose its fidelity in prior learned knowledge to a certain degree after learning knowledge from new tasks. As Fig.5 (c) shows, compared to other methods, the steering control of Non-LL is slightly unstable when turning, while by utilizing life-long learning methods LL-ME and LL-no-ME can guarantee a smoother steering control.

## V. CONCLUSION AND FUTURE DIRECTIONS

In this paper, a life-long adaptive path tracking policy learning method for autonomous vehicles that can adapt and evolve with continual multi-task knowledge is proposed. A memory evaluation method is proposed to optimize memory structure for better life-long policy learning. And a novel model-free path tracking policy model is proposed to learn vehicle dynamic responses and corresponding control action simultaneously from historical driving experience is proposed. Experiments are conducted within the simulated environment using high fidelity vehicle dynamic model in curvy roads. Experimental results show that the proposed method is effective in adapting and evolving with continual task knowledge learned without forgetting knowledge from prior tasks. Besides, the proposed method is capable of adapting to more difficult scenarios with knowledge from task knowledge accumulated from easier scenarios. Moreover, compared with two baseline methods, which are pure pursuit and model predictive control, the learned policy can achieve the minimum tracking error after evolving, which demonstrates the effectiveness of the proposed method in learning realistic vehicle dynamic responses and corresponding control policy.

The ultimate goal of this research is to apply the proposed life-long learning-based policy in real applications, where the policy can perform, adapt, and evolve continually. But it is worth noting that there are still some issues to be taken care of before applying the life-long learning-based method in real applications, which are also some interesting future directions. Firstly, as shown in our results, the policy will have to be pre-trained in a certain number of tasks to guarantee a basic performance in most scenarios before performing and continually evolving in new scenarios, where the number of tasks needed and corresponding evaluation methods should be further discussed. Secondly, although a simple preview point representation can already achieve very good performance in this work, richer representations should be considered to further improve the policy performance. Thirdly, environmental changes should be considered for adapting to more complex environments, for instance, off-road environments.

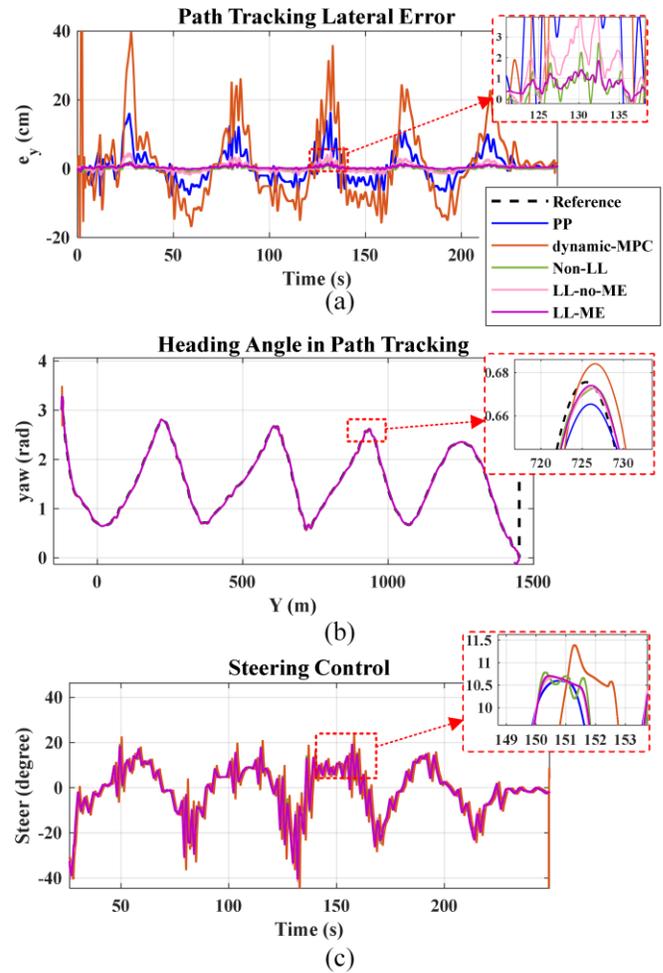

Figure 5. Experimental results of path tracking task in test scenario S1 using different policy, (a) the lateral tracking error, (b) vehicle heading angle and reference heading angle, (c) steering control output of different policy.